\documentclass[letterpaper, 10 pt, conference]{ieeeconf}  

\IEEEoverridecommandlockouts                              

\usepackage{graphicx}
\usepackage{cite}
\usepackage{picinpar}
\usepackage{amsmath}
\usepackage{url}
\usepackage{flushend}
\usepackage{colortbl}
\usepackage{soul}
\usepackage{multirow}
\usepackage{pifont}
\usepackage{color}
\usepackage{alltt}
\usepackage{enumerate}
\usepackage{siunitx}
\usepackage{epstopdf}
\usepackage{pbox}
\usepackage{subfigure}
\usepackage{textcomp}

\usepackage{amsmath}
\usepackage{amssymb}
\usepackage{amsfonts}
\usepackage{amstext}
\usepackage{amsthm}

\title{\LARGE \bf
Depth map estimation methodology for detecting free-obstacle navigation areas}

\author{Sergio Trejo, Karla Martinez and Gerardo Flores \thanks{Sergio Trejo, Karla Martinez and Gerardo Flores are with the Perception and Robotics Laboratory, Centro de Investigaciones en \'{O}ptica, Le\'{o}n, Guanajuato, Mexico, 37150. (email: sergiotrejo@cio.mx, daniesparza@cio.mx, gflores@cio.mx).}
\thanks{This work was supported in part by the FORDECYT-CONACYT under grant 292399, by the Laboratorio Nacional de \'Optica de la Visi\'on of the National Council of Science and Technology in Mexico (CONACYT) agreement 293411 and by the FORDECYT-CONACYT with the project 296737 Consortium in Artificial Intelligence.}
}

\begin{document}

\maketitle
\thispagestyle{empty}
\pagestyle{empty}

\begin{abstract}
This paper presents a vision-based methodology which makes use of a stereo camera rig and a one dimension LiDAR to estimate free obstacle areas for quadrotor navigation. The presented approach fuses information provided by a depth map from a stereo camera rig, and the sensing distance of the 1D-LiDAR. Once the depth map is filtered with a Weighted Least Squares filter (WLS), the information is fused through a Kalman filter algorithm. To determine if there is a free space large enough for the quadrotor to pass through, our approach marks an area inside the disparity map by using the Kalman Filter output information. The whole process is implemented in an embedded computer Jetson TX2 and coded in the Robotic Operating System (ROS). Experiments demonstrate the effectiveness of our approach.
\end{abstract}
%
\section{Introduction}
Recently, navigation of mobile robots in unknown environments has been an area of interest for researchers \cite{flores_vision_2014}, due to the increasing number of applications and the necessity of maneuver autonomously with efficiency. An important research issue is the obstacle and object detection by using vision techniques \cite{flores_vision_2013}. It has been developed in a number of methods and algorithms, among the most common methods we can mention those based on sensors like 3D-LiDAR, RGB-D cameras, monocular cameras, stereo cameras, among others \cite{zhou_vision-based_2013}. Each approach has its strengths and weaknesses, and several algorithms have been developed in the last few years in order to reduce the depth estimation error. In this paper, we present a simple approach to ameliorate the depth estimation given by a stereo camera rig together with a 1D-LiDAR. With this information is defined a window in which a quadrotor can be navigate freely.
\begin{figure}[t]
   \centering
    \includegraphics[width=\columnwidth]{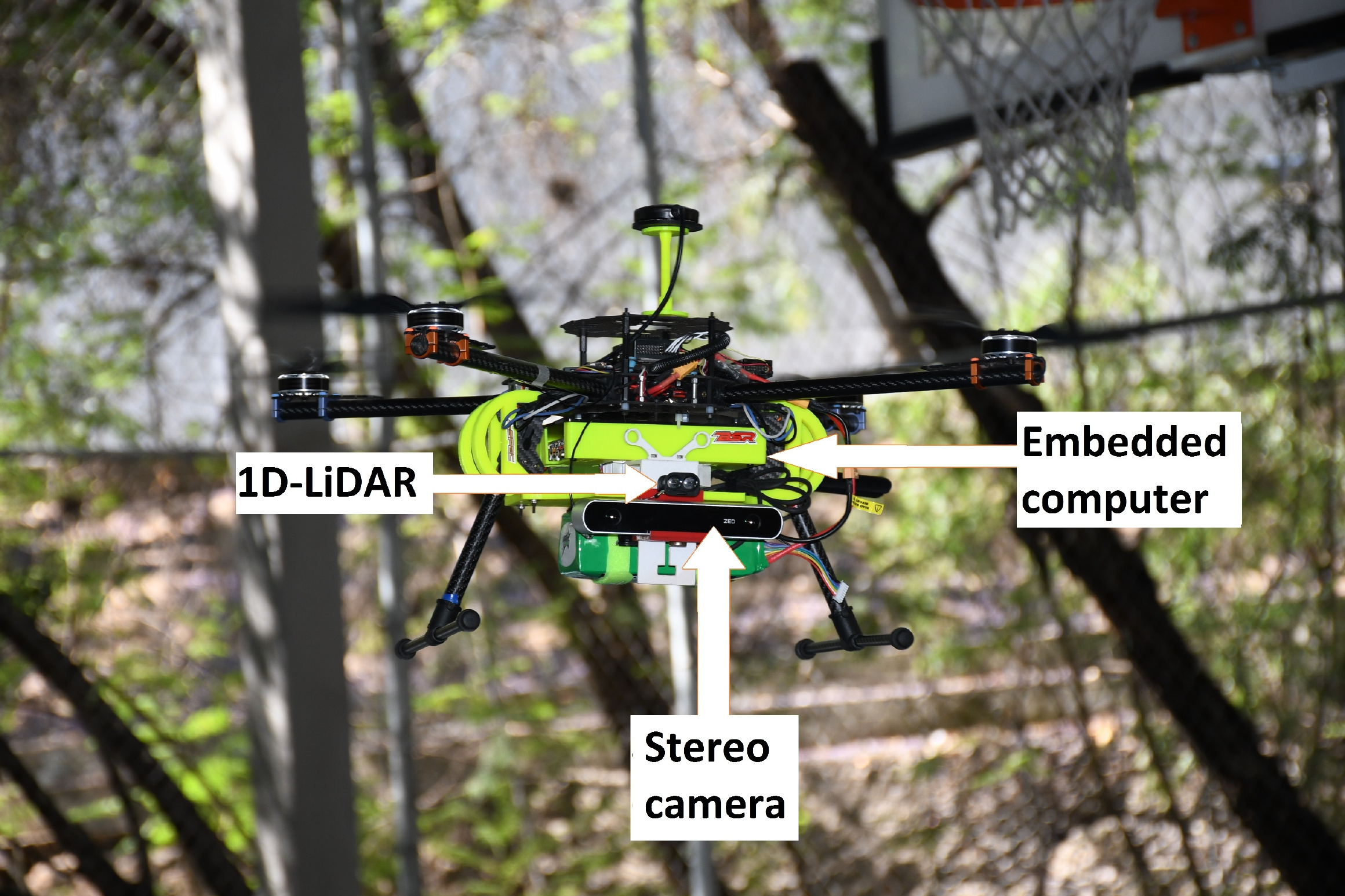}
    \caption{Quadrotor UAV used in this paper endowed with a Jetson TX2, stereo camera rig (ZED camera), and Lidar Lite sensor.}
    \label{drone}
    \end{figure}
\subsection{Previous work}
The first issue to solve is to choose an algorithm of stereo matching, which is crucial to obtain a good disparity map. In the website Middlebury Stereo Evaluation - Version 2 \cite{middlebury2} there is a list of more than 150 stereo matching algorithms ranked according to the average percent of bad pixels, obtained from the relation between the computed disparity map and the ground truth. However, the best reference and to compare stereo matching algorithms is the version 3 website of the aforementioned evaluation site \cite{middlebury3} which is based on the paper of D. Scharstein \cite{taxonomy} where several parameters of stereo matching are defined with the purpose of comparison. In this list we find the latest stereo matching algorithms and the most accurate are predominantly performed by neural networks or superpixels methods or a mixture of them. Among the more accurate methods is \cite{lincheng}, who proposes an algorithm based on superpixels labeling of an image and then applying a bilayer matching cost where a neural network compare similarity between layers. This kind of approach reduces the disparity map noise but the computation time increases significantly. Meanwhile H. Hirschmüller \cite{hirschmuller} proposes the Semi-Global Block Matching (SGBM) method and it works faster but with less precission. 

Regarding algorithms for obstacle avoidance, for instance A. Stanoev et al. \cite{stanoev} establish a threshold in the depth map where the close objects are white and labeled as obstacles and the farther ones are black and then ignored; if the robot moves quickly, the threshold decreases. In some cases it is necessary to differentiate obstacles over a flat surface, in this case it is useful to implement V-disparity maps \cite{vdisp}  which is a function of the disparity map, that accumulates the disparities of the horizontal line into the v-disparity function, where the abscissa corresponds to the number of disparities. This approach can be used in vehicle navigation on a road. B. Lopez \cite{brett} proposes a perception and planning approach that significantly reduces the computation time using instantaneous perception for obstacle avoidance. Aman \cite{aman} proposes a methodology to fuse ultrasonic sensor measurement and depth map from Kinect sensor. M. ki et al \cite{ki} propose a framework which implements a stereo camera and a 2D-LiDAR on an UAV, however the sensor is the only obstacle detector, and the camera is just used for monitoring. H. Song \cite{song} proposes the fuse of RGBD and 2D-LiDAR for tracking purposes. Roopa et al. \cite{roopa} fuse images using Kalman Filter (KF) to get more information about the localization of a target, this approach is applied to different cameras and different localization. In \cite{sharma} the authors fuse with a KF three distance sensors in order to obtain the distance and orientation with respect to a wall.  In \cite{park} K. Park et al. presents a high-precision depth map using a high cost 3D-LiDAR, however, cost of implementation is considerably higher than the approach presented in this paper.
\subsection{Main contribution}
One of the key points in UAV autonomous navigation is the obstacle avoidance problem. In this work, we address the problem of identifying free navigation areas instead of detecting a particular obstacle. We have chosen such an approach due to the high complexity in determining a broad class of objects when we deal with object classifier approach \cite{gomez-balderas_tracking_2013}, \cite{zhou_real-time_2015}. For that aim, we use the information of two low-cost devices: a stereo camera rig and a 1D-LiDAR. With the stereo camera we estimate a disparity map, then we measure the distance in front of the quadrotor with a 1D-LiDAR. Then, both data are fused in a KF to obtain a better estimation of the distance between the front of the quadrotor and a predefined area where the UAV can navigate as long as such a distance is free of obstacles.
\subsection{Organization of the rest of the paper}
In Section \ref{sec:problem} we present in detail the problem formulation. In Section \ref{sec:methods} the proposed methodology is described, in which the system overview, depth map estimation algorithm and KF are presented. Section \ref{sec:exp} shows experimental results. Finally, at Section \ref{sec:conc} we present some concluding remarks and future research.
\section{Problem formulation}\label{sec:problem}
\subsubsection{Problem}

There are several methods for disparity map computation presented in the literature, however, many of them throw a disparity map with a great quantity of noise, this noise could be interpreted as obstacles. 

To deal with this problem the most popular solution is the use of expensive computation, which slows down the disparity map generation. Other faster algorithms can eliminate the noise but can be imprecise and inaccurate.   

To obtain a precise measurement one of the most popular solution is the use of 3D-LiDAR, however this class of sensors are expensive and heavy, making them an option not easily available. Other option are the RGB-D sensors, but they do not work properly under daylight environment.
\subsubsection{Solution}
\begin{figure}[t]
    \centering
    \subfigure[Without obstacle.]{\includegraphics[width=40mm]{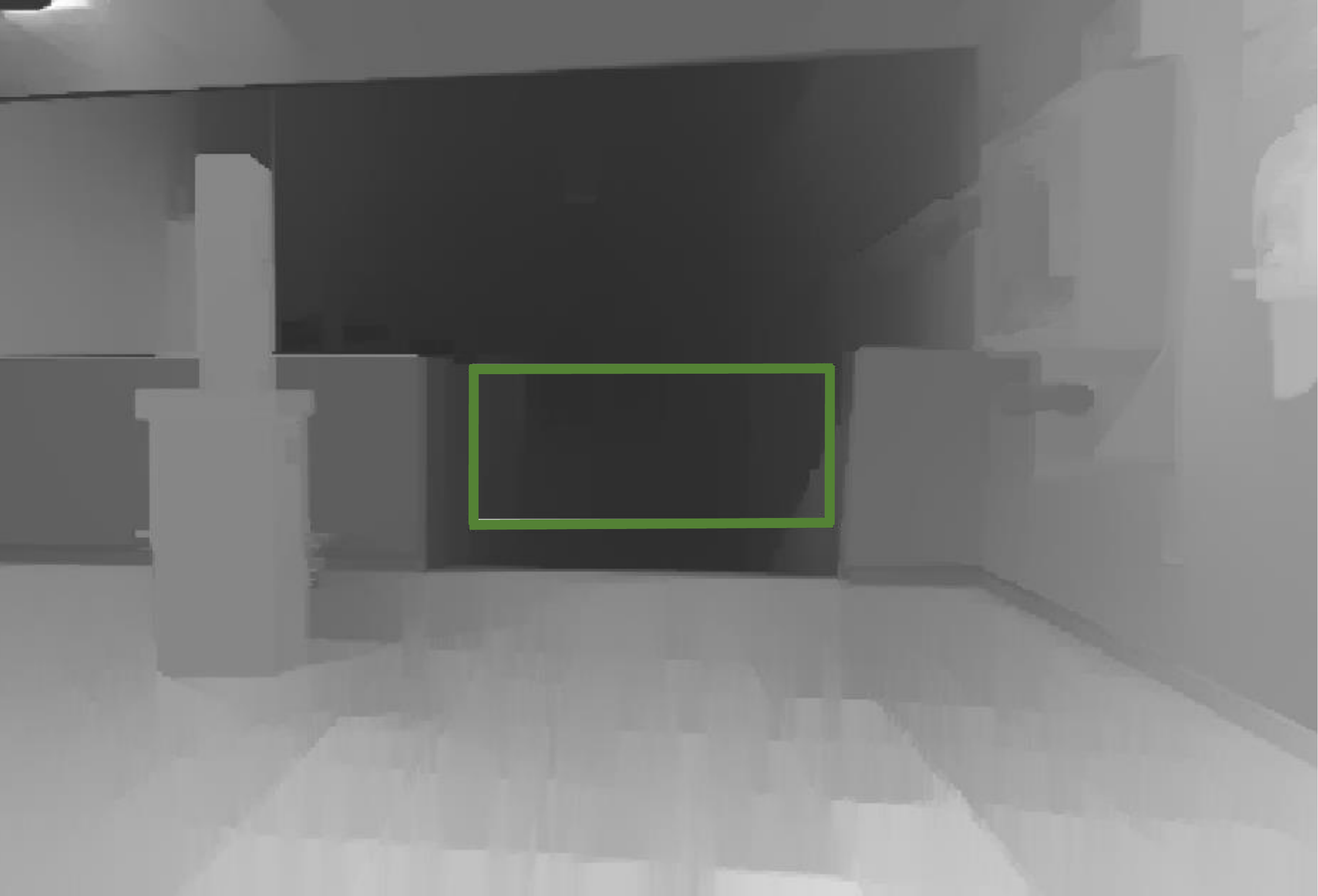}}
    \subfigure[With obstacle.]{\includegraphics[width=40mm]{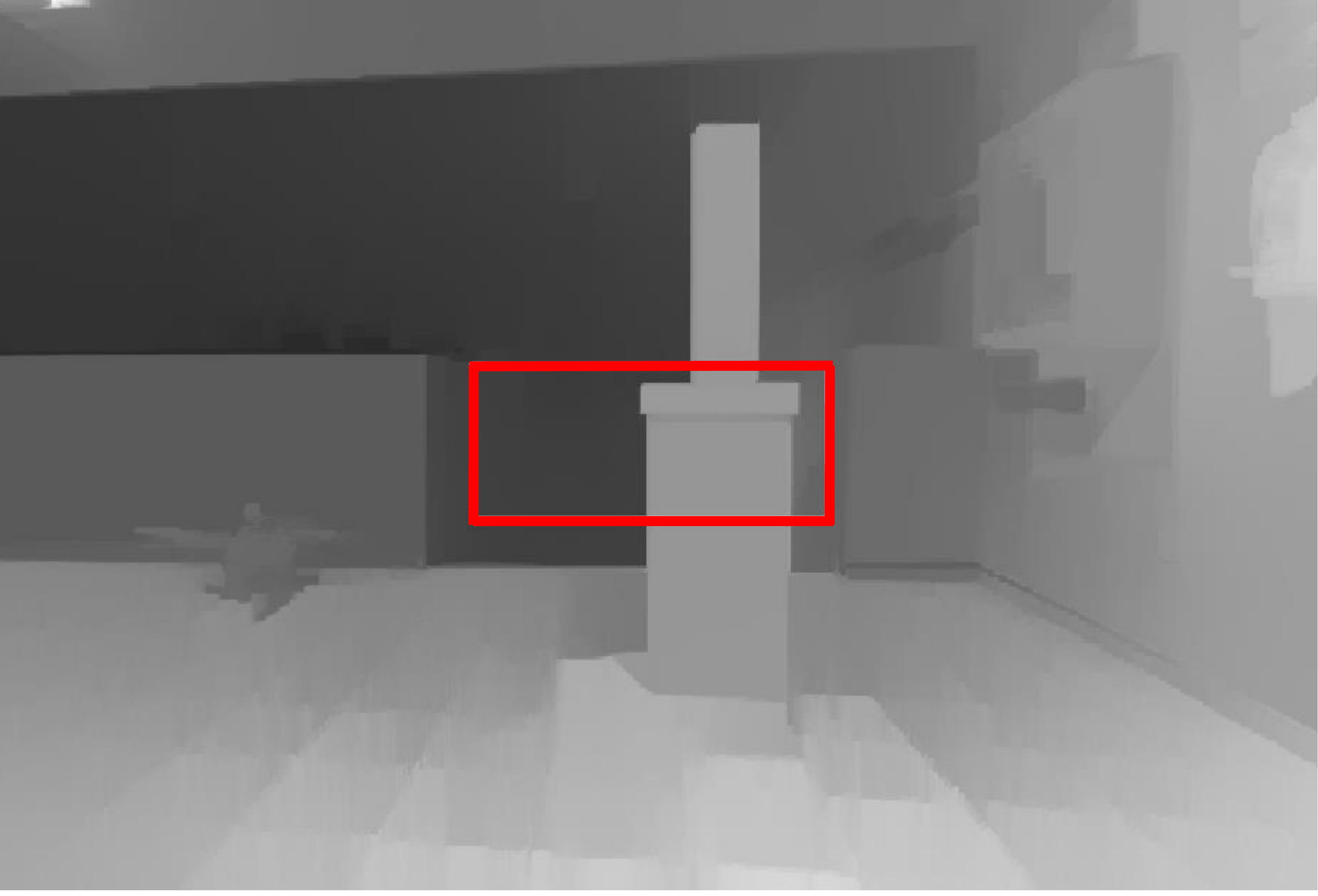}}
    \caption{The (a) image shows a free area where the UAV is able to navigate. Fig. (b) represents an obstacle inside the rectangle, which does not allow the UAV to navigate. Observe that the rectangle represents the quadrotor's size in which the navigation path must be free, therefore the goal is to determine if this area is blocked or not by a potential obstacle.}
    \label{obstacle}
\end{figure}

We need to ameliorate the depth map using only: a) a stereo camera rig with disparity estimation algorithms free of noise; and b) a simple and low-cost 1D-LiDAR sensor of one dimension, (not the popular Hokuyo, but the simple one with a cost of around 100 USD). With that aim, we propose to fuse the information from both devices in a KF. Also, for safe navigation we compute a safe distance ($4$ meters) in which the quadrotor can safely navigate. Then the quadrotor must determine if there is or not any object that can block the UAV navigation path. For that, it is determined a window considering real quadrotor dimensions. Such a window is depicted in the scene captured by the stereo camera rig, a picture of this idea can be seen at Fig. \ref{obstacle} where the rectangle represents the UAV size to a distance from the camera equal to $4$ meters. As it is shown in the Fig. \ref{obstacle}, when an obstacle is present, a red rectangle appears, while in the opposite case (free obstacle path) the rectangle is green. The 1D-LiDAR is pointing in the rectangle centroid, and hence a depth estimate is computed in a Kalman Filter algorithm.
%
%
\section{Methodology}\label{sec:methods}
In this section we present the system overview, the information acquisition processes for 1D-LiDAR and stereo camera rig. Also, the implemented KF is presented.
\subsection{System overview}
The system is conformed by sensors, microcomputers, software and a quadrotor UAV. In this part we begin by describing the sensors, particularly the stereo camera rig and the 1D-LiDAR. How the provided information by these sensors is processed and interpreted is presented next. Later, we present results about implementation. 

\begin{figure}[t]
    \centering
    \includegraphics[width=\columnwidth,angle=-90]{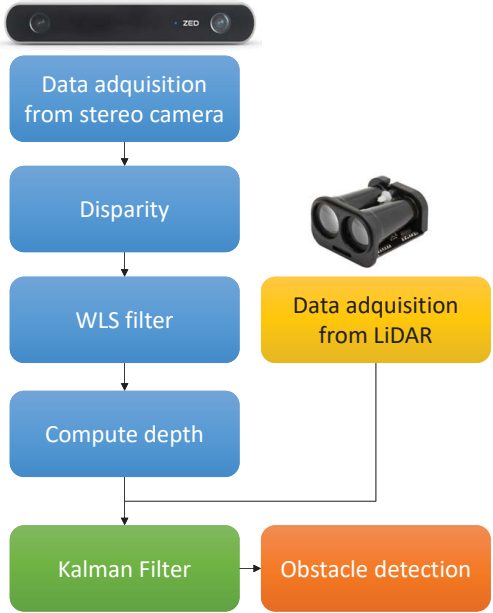}
    \caption{System overview. Blue blocks: Camera processes. Yellow block: 1D-LiDAR processes. At bottom, both information are fused in a Kalman Filter.}
    \label{Overview}
\end{figure}

First, consider the pair of images provided by stereo camera. For each frame two measurements are captured and then processed according to the system overview shown in Fig. \ref{Overview}. The KF have the aforementioned measurements as inputs which correspond to the disparity map and the distance acquired by the 1D-LiDAR. After that, inside the depth map, it is computed a rectangle with the height and width corresponding to the dimensions of our UAV at a determined distance in front the UAV. The dimensions of our UAV is about of $120$ cm of width and $40$ cm of height. The center of the image captured corresponds to the same physical point that the 1D-LiDAR measures as distance. Then, the rectangle is an area that must be free of any obstacles to avoid any possible collision. If we are able to measure the distance with accuracy, the size of quadrotor can be projected ahead and determinate if the UAV can access freely across of the rectangle.
%
\subsection{Depth map estimation} \label{sub:depth}
There are several algorithms under research for depth map estimation based on stereo cameras, some of them are faster than others, but the faster the algorithm the less accuracy of computing of the depth map. It is important to mention that in the quadrotor navigation, the algorithm efficiency is crucial, since all the calculations are performed onboard the embeded computer which has limited capacities. For the purpose of this paper we considered two important parameters for disparity map estimation: the quality of the depth map and the computation time. According to the Middlebury Stereo Evaluation \cite{middlebury3}, Semi-Global Block Matching \cite{sgbm} (SGBM) algorithm has an absolute average error of $14.3$ pixels and $0.68$ sec/megapixels which can be implemented with OpenCV \cite{hirschmuller} easily. The absolute average error is the absolute difference of disparity between the computed disparity map and the ground truth map, whereas the time of computation is the time it takes a one million of pixels disparity map to be computed. The SGBM algorithm is based on smoothness constrains applied on pixel-wise matching which removes outliers.

The approach used to achieve the disparity map from camera is performed through OpenCV libraries, which is based on Hirschm\"uller algorithm with some modifications added by OpenCV developers. Once the disparity is obtained we used a Weighted Least Squares filter \cite{cv_wls} (WLS) to remove holes due to half-occlusions, in other words WLS filter fills the image to get a uniform segmentation between layers. 
Depth is computed from the following equation
\begin{equation}\label{depth}
    \textrm{D} = \frac{f B}{d}
\end{equation}
where $D$ is the depth, \textit{f} is the focal distance in mm, \textit{B} is the baseline between both cameras in pixels, \textit{d} is the disparity got previously in pixels.In \eqref{depth} the depth depends on focal distance and camera baseline, which are intrinsic parameters of camera and they are unique values for each camera rig. Then, depth depends on disparity and on the camera calibration matrix. According to epipolar geometry \cite{hartley} the disparity is the difference in pixels between the projection of a 3D point in the right camera and the same 3D point projected on the left cameras, as the images were overlapped. If there is an erroneous disparity or calibration, the depth will be erroneous. For that reason, parameters are modified to achieve a better matching between two images obtained from the camera. The application of the WLS filter help to remove the NaN (Not a number) and Inf (infinity) values. WLS filter was implemented as a fast global smoother but running on CPU; the implementation on GPU is left for later.
\subsection{Kalman Filter implementation}
The quadrotor position is modeled as decoupled linear stochastic system based on the Newton Second Law \cite{arreola_improvement_2018}. For instance, the $x$ quadrotor position dynamics is modeled as
\begin{equation}\label{newton}
        \ddot{x}=a u  + w_{t}
\end{equation}
where $u$ is the acceleration given by the control command and $w_{t}$ is the model uncertainty. With no loss of generality $a=1$ in \eqref{newton}, where it represents a system parameter. The state space model of \eqref{newton} with the change of variables $x=x_1$ and $\dot{x} = x_2$ and with available outputs is given by
\begin{equation}
			\dot{x}(t)
				=\left[ \begin{array}{cc}
			0 & 1 \cr 0 & 0
				\end{array} \right] 
			x(t)
				 + \left[ \begin{array}{c}
			0\cr 1
				\end{array} \right] u + w_{t} \label{contstatespace}
\end{equation}
\begin{equation}
			y(t)
				=\left[ \begin{array}{cc}
			1 & 0 \cr
			1 & 0 \end{array} \right] 
			x(t) + v_{t} \label{contmeas}
\end{equation}
where the outputs are given by the sensor measurements, i.e. the 1D-LiDAR and the disparity map; the $(w_{t}, v_{t})$ are the process and measurement noise, respectively. Both variables are assumed as white noise and has zero mean. In discrete time the state space model and measurement model are given as follows
\begin{equation}\label{state}
{x}_k=\mathbf{A}x_{k-1}+\mathbf{B}u_k+w_{k} 
\end{equation}
\begin{equation}\label{measure}
y_{k}=\mathbf{C}x_{k}+v_{k} 
\end{equation}
where the state transition matrix is 
\begin{equation}
    \mathbf{A}=\left[\begin{array}{cc}
    1 & \Delta T\cr 0 & 1
    \end{array}\right]\label{a}
\end{equation}
in which $\Delta T$ is the sampling time, which is the time between each frame computed, $w_k$ is the noise of process, $v_k$ is the measurement noise. The matrix $B$ is the control input matrix. Since we have two sensors that measure the same variable, i.e. distance, the matrix $C$ can be expressed as
\begin{equation}
\mathbf{C}=\left[
    \begin{array}{cc}
     1 & 0\cr 
    1 & 0\cr
    \end{array}\right].
\end{equation}
The covariance matrix $Q = E[v_k v_k^T]$ of measurement vector $y_k$ can be expressed as
\begin{equation}
Q =\left[\begin{array}{cc} 
    \sigma_{z} & 0 \cr 
    0 & \sigma_{l} \end{array}\right]
    \label{covariance}
\end{equation}
where $\sigma_z$ and $\sigma_l$ are the variances of the depth map estimation and 1D-LiDAR sensor, respectively. This measurement noise matrix is diagonal, since we suppose the acquisition data from sensors is independent between both \cite{hossein}. The covariance matrix of the process is represented as $2\times2$ matrix, which seems to equation \eqref{covariance}, but the process covariance values are considered smaller than the measurement covariance matrix because as we stated, we assumed the measurement process when the quadrotor is static and modeled with the Second Law of Newton. To simplify this we assumed the process variances equal to zero.

%
The KF can be implemented in Robotic Operating System (ROS) with the contribution of D. Ratasich et al \cite{ratasich}. ROS is a set of libraries and tools that helps to build robot applications. In there, can be created packages composed of nodes. This nodes can be modular drivers or algorithms that runs together with a purpose. The aforementioned work can be used for generic sensor fusion purposes that can be configured according to the requirements of the system, in this manner, it is just necessary to to define the state space matrices including the process noise matrices and the initial state.

\begin{figure}[t]
 \centering
    \subfigure[]{\includegraphics[width=\columnwidth]{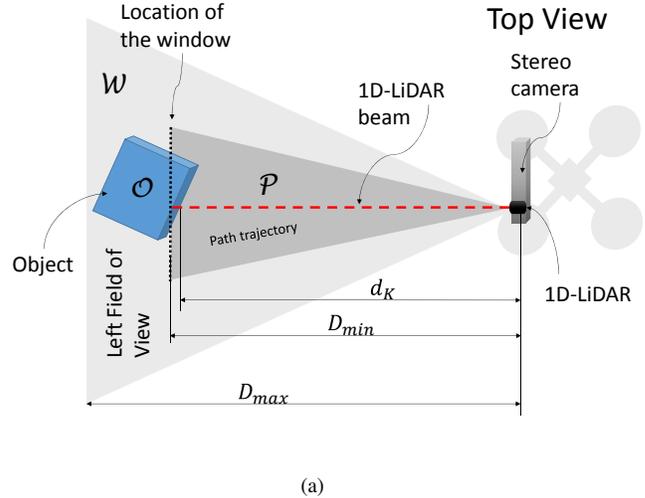}}
\subfigure[]{\includegraphics[width=\columnwidth]{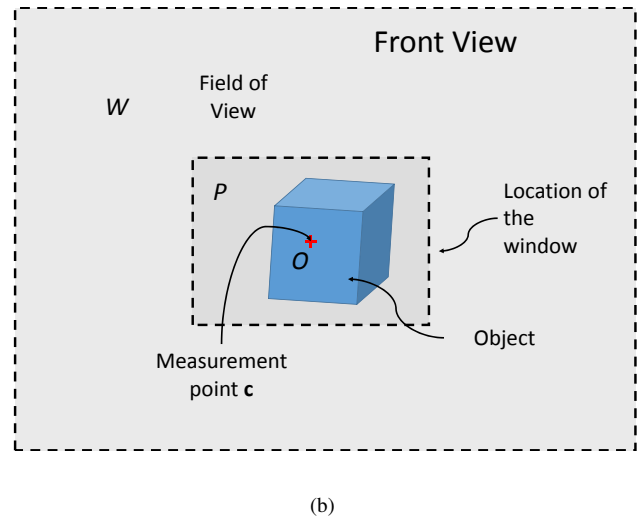}}
 \caption{Diagram representing the main elements of our approach. Part a) is the top view of the obstacle detection; and b) is the view seeing by the camera mounted on the drone.}
 \label{obstaclescheme}
\end{figure}

\subsection{Obstacle detection} 
Let $\mathcal{W}\subset \mathbb{R}^3$ be the set of all points in a world coordinate system that are mapped inside the field of view (FOV) of the stereo camera. The FOV can include objects, some of them can be considered as obstacles if they are in the path space $\mathcal{P}\subset \mathcal{W}$. We represent obstacles by the set $\mathcal{O}$. Then, if the following inequality holds
\begin{equation*}
    \mathcal{O} \cap \mathcal{P} \neq \emptyset 
\end{equation*}
we know that the path is free to navigate. Let consider the filtered depth map explained in the previous section. There is a maximum reach distance of the depth map, named $D_{max}$, please refer to Fig. \ref{obstaclescheme} a). Also, there is a \textit{window} in which the UAV can pass trough without any obstruction shown at Fig. \ref{obstaclescheme} b) and at Fig. \ref{exper_ext}. Such window is at a distance $D_{min}$ in the $z$-coordinate from the stereo camera, please see Fig. \ref{obstaclescheme} a).

On the other hand, we know from epipolar geometry \cite{hartley}, that it exists a function that maps 3D points to 2D points, called 3D projection. Every 3D point and set in the field of view has a projection to the 2D plane in the left camera sensor (by convention) of the stereo rig, then

\begin{equation*}
    \begin{array}{cc}
			\mathbf{W}_i \mapsto \mathbf{w}_i \cr 
			\mathbf{P}_i \mapsto \mathbf{p}_i  \cr
			\mathbf{O}_i \mapsto \mathbf{o}_i \cr 
    \end{array}
\end{equation*}
where $\mathbf{W}_i = (x_i,y_i,z_i) \in \mathcal{W}$ represents a point in the FOV; $\mathbf{P}_i\in \mathcal{P}$ is a point inside the path of the UAV; $\mathbf{O}_i\in \mathcal{O}$ is a point of a physical object; $\mathbf{w}_i \in W \in \mathbb{R}^2$ is a point of the FOV projected in the camera sensor; $\mathbf{p}_i \in P \in \mathbb{R}^2$ is a projected point of a path in the camera sensor, in other words, it is the window projection in the camera; and $\mathbf{o}_i \in O \in \mathbb{R}^2$ is a projected point of an obstacle, see Fig. \ref{obstaclescheme}b . All the above for $i=\text{\{ 0,1,2,3,...\}}$.

We use the depth function \eqref{depth} by means of the disparity between a point $\mathbf{p}_i$ seen from the right and the left camera sensors, in this way $D(\mathbf{p}_i)$ is the depth value in the position $(u_i,v_i)$ of the depth map. Every value of $(u_i,v_i)$ in the depth map represents the $z_i$ coordinate value of the point $\mathbf{P}_i$ that is projected to the image seen with respect to the left camera. Since our approach only considers displacements in the $z$ coordinate (forward direction) with respect to the UAV, we propose to use the window projection as a rectangular region centered inside the depth map. Such a region represents a free area in which the UAV can navigate. Then, $\mathbf{p}_i \in P \in \mathbb{R}^2$ represents the set of all points contained inside such a rectangular region, i.e. $P\subset W$. Inside the window region, we highlight the window centroid given by $\mathbf{c}\in P$ which is ideally the same point mapped to $\mathbf{P}_i$ that the 1D-LiDAR is pointing at. At $\mathbf{c}$, we get two measurements: $d_L$ which is the distance measured by the 1D-LiDAR, and $D_c=D(\mathbf{c})$ which is the distance obtained from the depth map. When both measurements are fused together we obtain $d_K$, which is compared with $d_{min}$; if $d_K < d_{min}$ then the point $\mathbf{c} \in \mathcal{O}$, i.e. it is part of an obstacle.


\section{Experimental validation}\label{sec:exp}
In this section we present the experimental platform and the obtained results with the approach presented in previous sections.
\subsection{Experimental setup}
The quadrotor UAV is developed at the LAB and is based on a Tarot XS690 quadrotor frame. As stereo camera rig we have mounted a ZED camera developed by Stereolabs, which is attached in the upper part of the quadrotor. This stereo rig has a specified range up to 40 meters and a number of third party support, among them, ROS and OpenCV. In these software elements: ROS and OpenCV we have coded our algorithm. Particularly we have used the ZED camera node provided by ROS, which gives the rectified right and left images that are required to compute the disparity. The disparity and autonomous operation is performed on-board with a Jetson TX2 development kit, a embedded computing board developed by Nvidia that stands out for being designed for machine learning applications. A laser distance sensor SW20/C \cite{lw20} is fixed between the left and right camera. Both cameras and sensor are positioning facing forward, in $x$ direction of the quadrotor. 

It is worth mentioning that ZED camera driver needs the Jetson computer to have installed JetPack 3.3 \cite{jetpack} which includes CUDA toolkit. In order to remove the disparity discontinuities by means of the WLS filter implementation, previously explained in Section \ref{sub:depth}, one needs to have OpenCV version 3.4.2 built together with the Extended Image Processing module (ximgproc), that is included in OpenCV Contrib repositories. The builder used for this purpose is the CMake. Nevertheless, WLS filter decreases significantly the frame rate of disparity map computing, so the node code in finally developed in C++, since it has a faster performance than Python in real-time image processing.
\begin{figure}[t]
    \centering
    \includegraphics[width=\columnwidth]{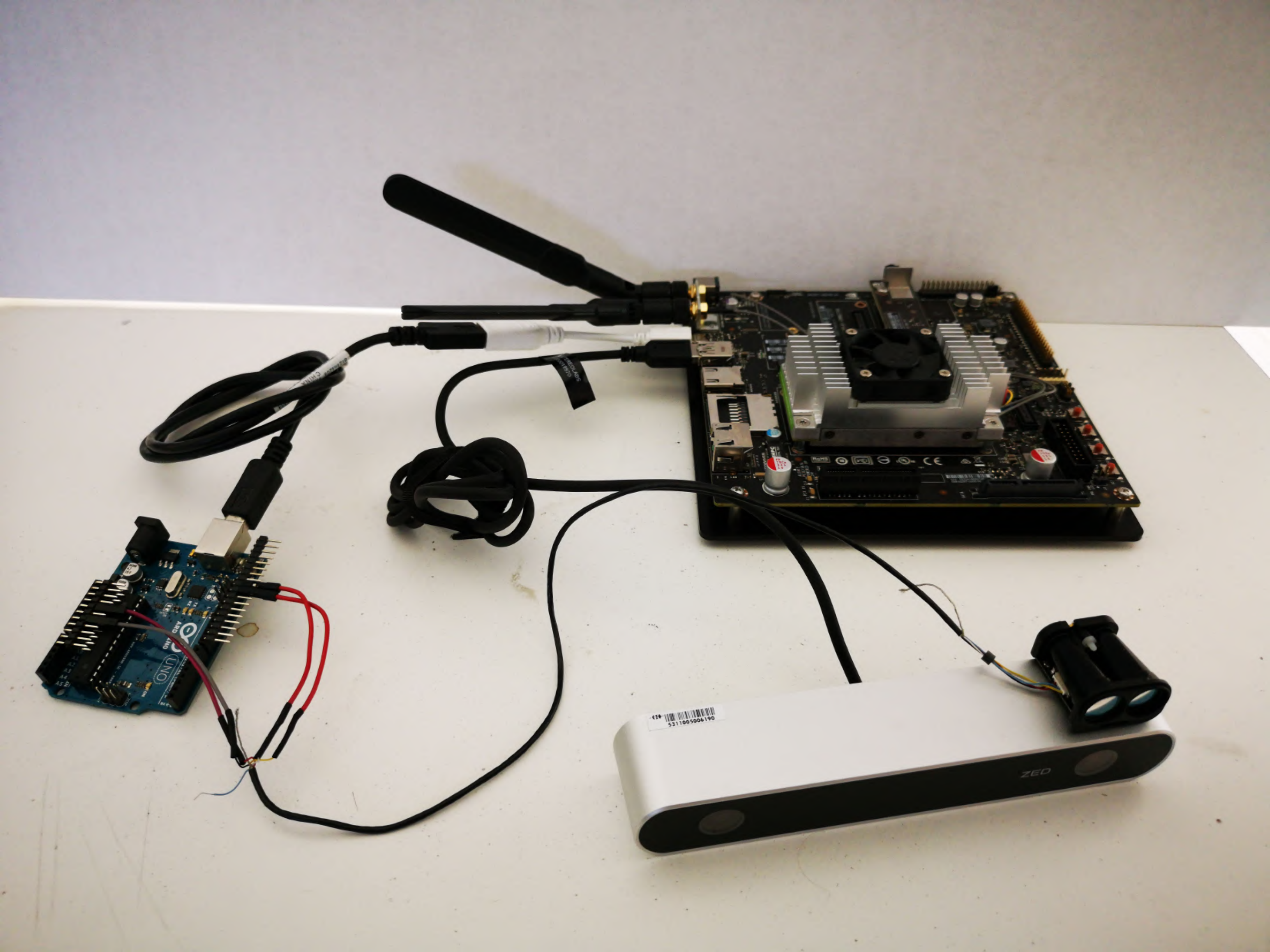}
    \caption{Components: ZED camera, 1D-LiDAR sensor, Jetson TX2 and Arduino UNO.}
    \label{zed}
\end{figure}


The sensor 1D-LiDAR device used has a specified detection range from 0.2 to 100 meters, with a resolution of 0.01m and an accuracy of 10cm, however, according to our experiments it has a minimum detection distance of 8cm, and its accuracy changes depending on the surface the laser beam is pointing at. When it is an opaque surface, its accuracy is much less than 10cm, but if the surface is more or less transparent, the distance accuracy is affected some times more than 10cm; for instance, when measuring the distance to a computer monitor, of measuring through a glass window, etc. The 1D-LiDAR position is near the left camera since the disparity and distance measuring must match with the same object to the extent as possible. All the components are shown at the Fig. \ref{zed}.

The data acquisition from laser distance sensor to the Jetson TX2 is performed with a code that reads data from $I^2C$ bus of an Arduino Uno microcontroller and then sent to Jetson TX2 via USB.  On the hand, in ROS environment the acquisition is performed with Lidar Lite ROS package \cite{roslidar}, although we have modified the code to our requirements.

%
Kalman filter is implemented with the ROS package created by Denise Ratasich et al. \cite{ratasich}, where we have modified the parameters accordingly to the model of our system, i.e. covariance matrices, inputs, ouputs, matrix dimensions and states. 
\begin{figure}[t]
 \centering
    \subfigure[Without WLS filter
    ]{\includegraphics[width=40mm]{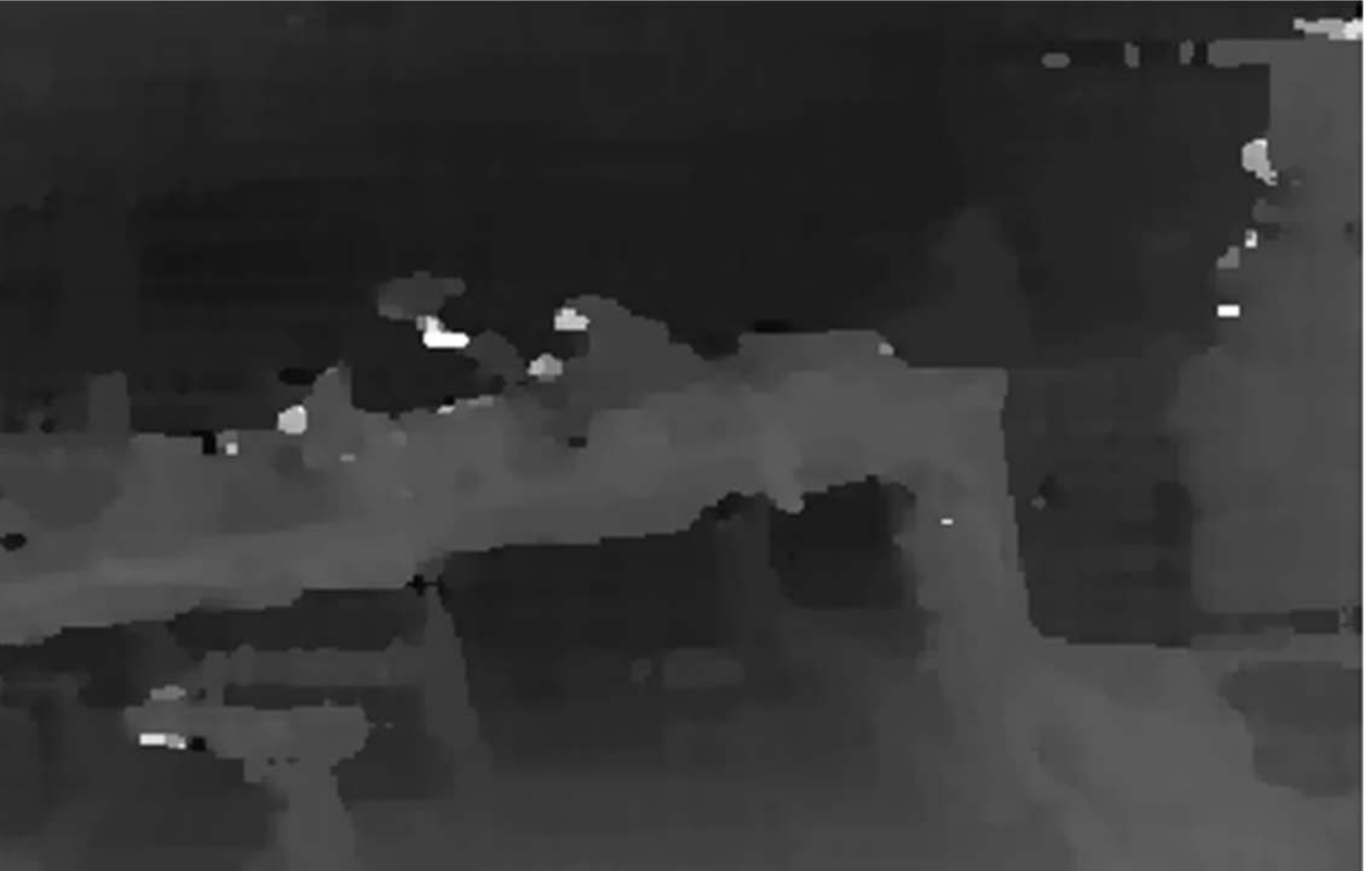}}
\subfigure[With WLS filter]{\includegraphics[width=40mm]{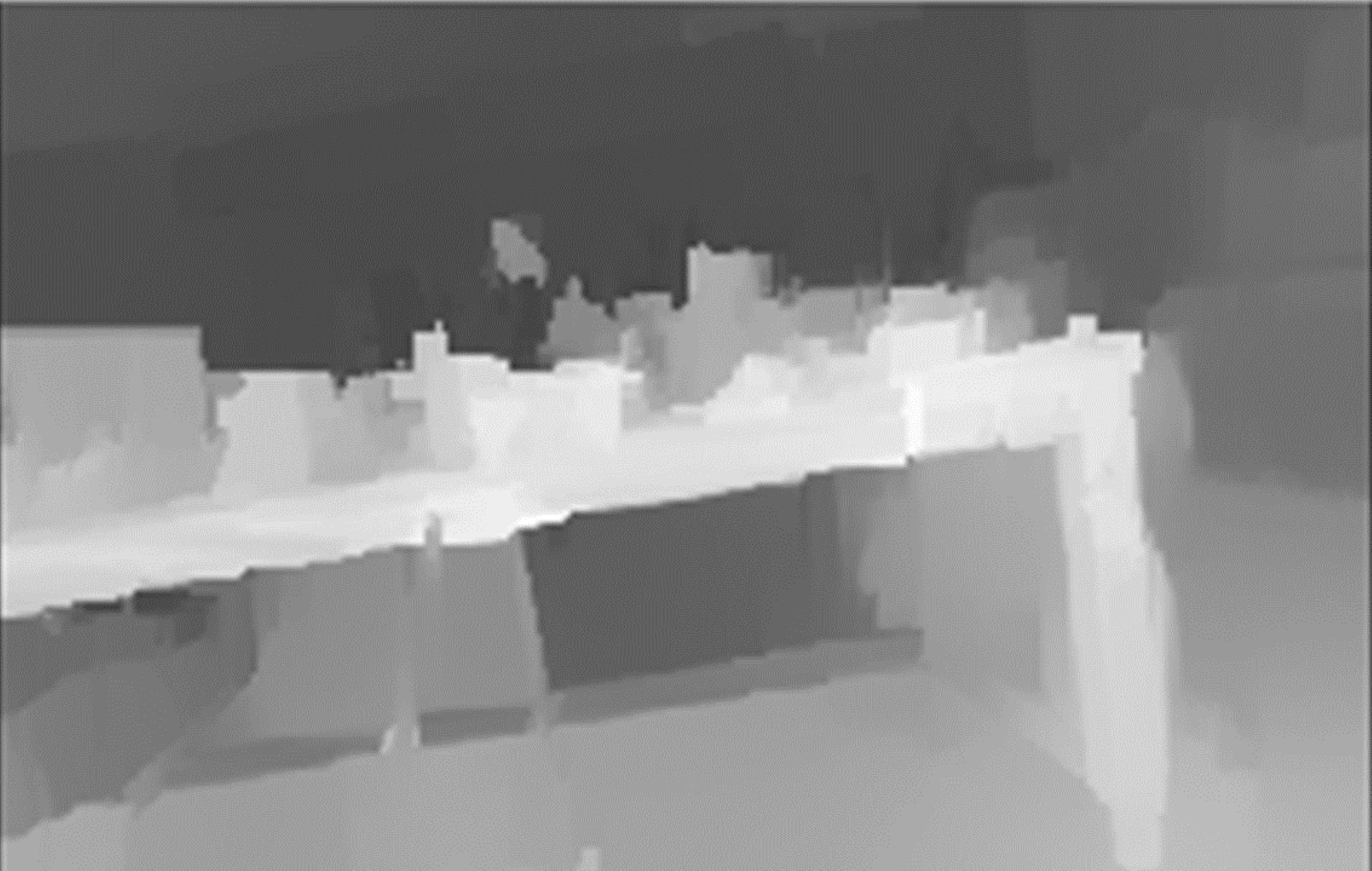}}
\subfigure[Without WLS filter]{\includegraphics[width=40mm]{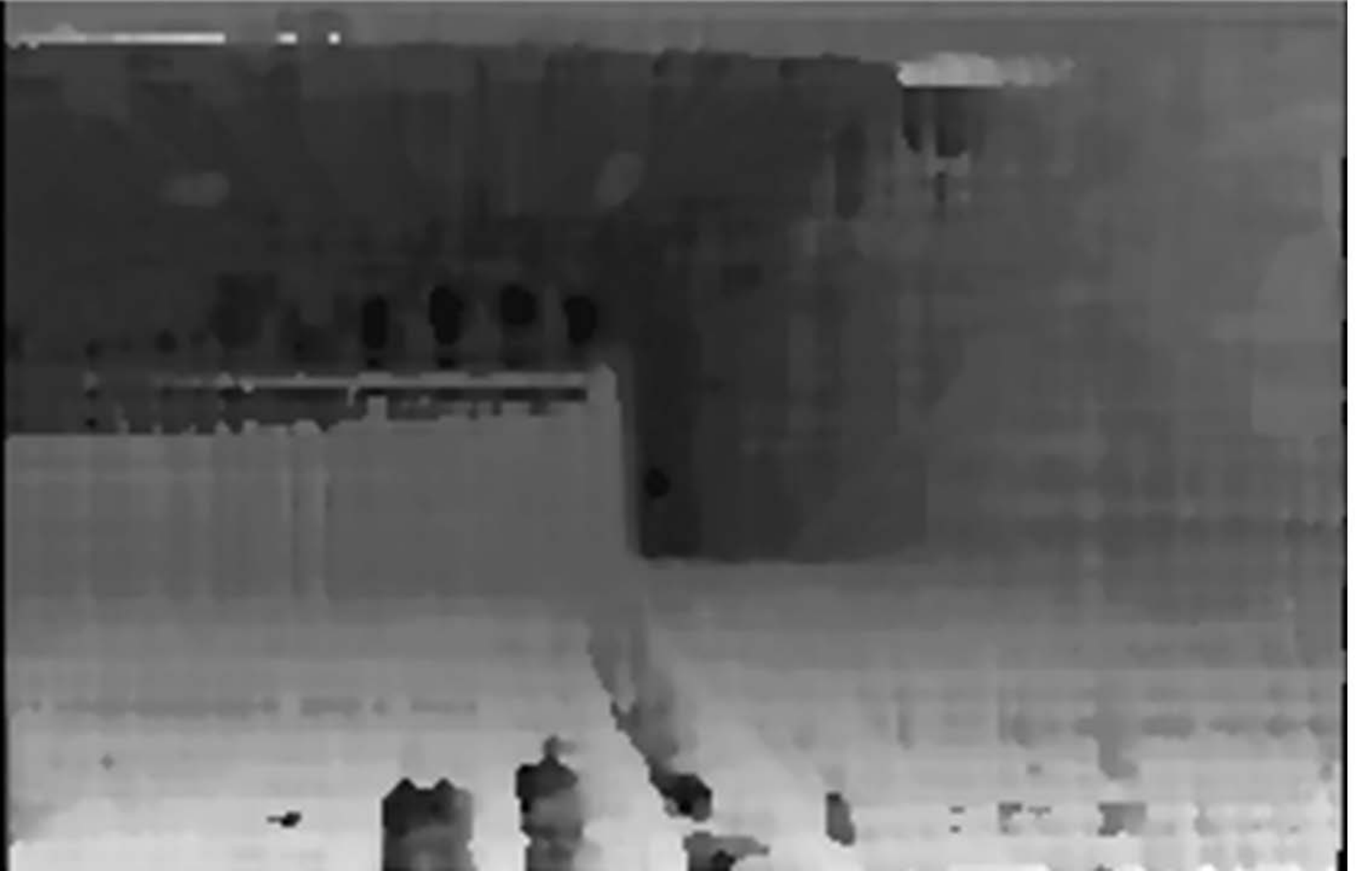}}
\subfigure[With WLS filter]{\includegraphics[width=40mm]{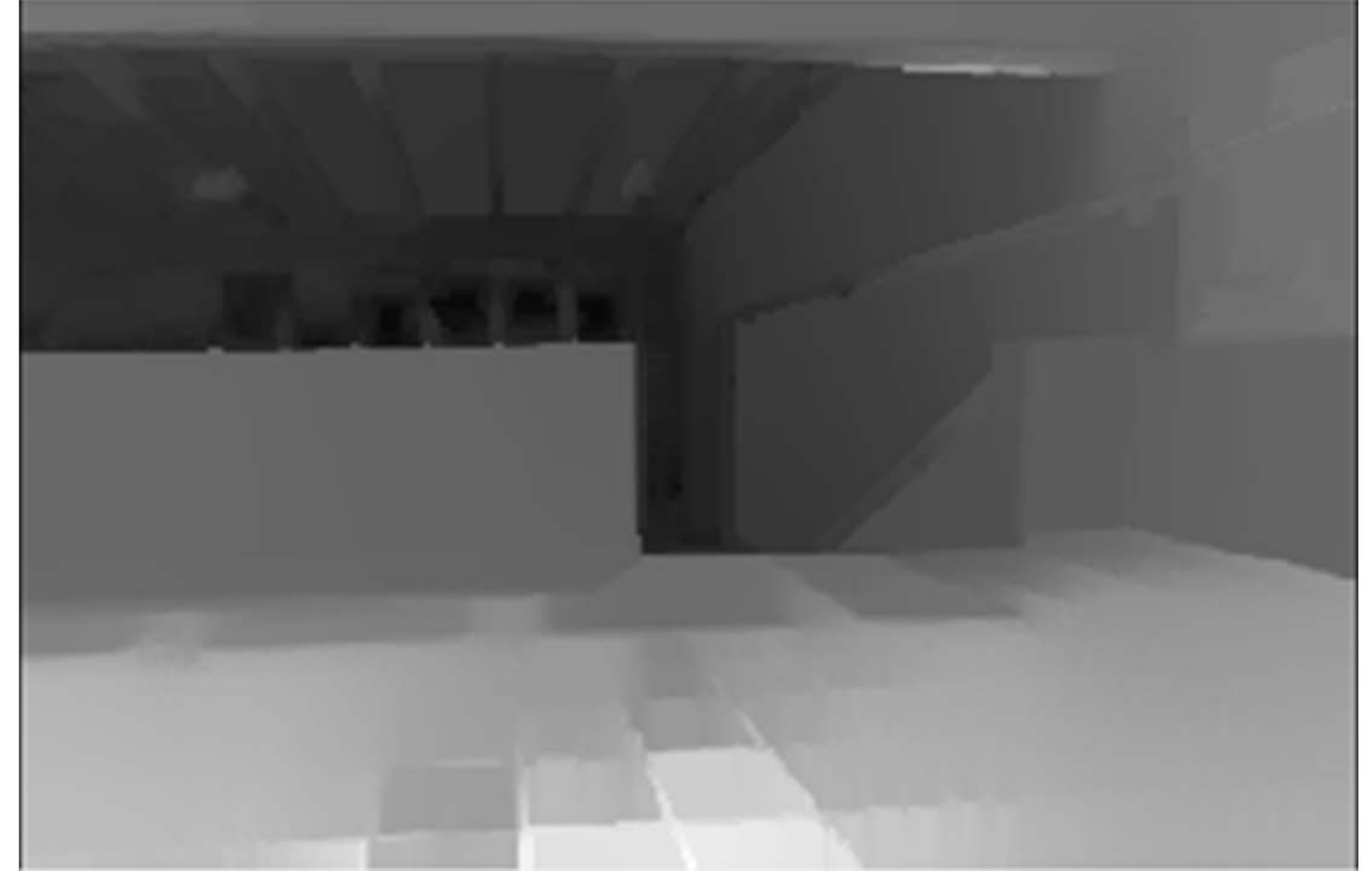}}
 \caption{Disparity with and without WLS filter.}
 \label{disparityfilter}
\end{figure}
\subsection{Results}
The covariance matrices were computed taking samples with the camera in a period of time whose center was pointing at the same point that laser sensor, and using the statistics covariance formula \cite{devore}
 \begin{equation*}
    \sigma ^ 2 = \frac{\sum(x_i-\bar{x})^2}{n-1} \label{variance}
\end{equation*}
where $n$ is the number of samples; $x_i$ is the measurement; and $\bar{x}$ is the expected value. We obtain both measurement noise. In this calculation, we fixed the stereo camera and the 1D-LiDAR. For stereo depth measurement $150$ samples was taken, pointing at a fixed object at a distance of $4$ meters. Therefore, We obtain $\sigma_z=0.0254800198$ corresponding to the disparity map. For laser distance sensor $150$ samples were taken simultaneously, obtaining $\sigma_l=0.0005798584$.

The results obtained with disparity algorithm was conducted with OpenCV using three processes to an image of 640x480 pixels: StereoSGBM function, WLS filter and normalization, and are shown in Fig. \ref{disparityfilter}, where a) and c) illustrate the disparity map without WLS filter, and b) and d) are the ones coming from the three processes. It can be observed that the non-filtered disparity map has some white and black spots, therefore, it generate bad measures of depth when they appear in the center of the map, which usually are caused by NaN, negative or infinity values. Nevertheless, the images with a complete disparity without white and black spots show smoother images with a better disparity, this means that the computation of depth map has less noise. 
\begin{figure}[t]
 \centering
 \subfigure[No obstacle inside window in RGB.]{\includegraphics[width=40mm]{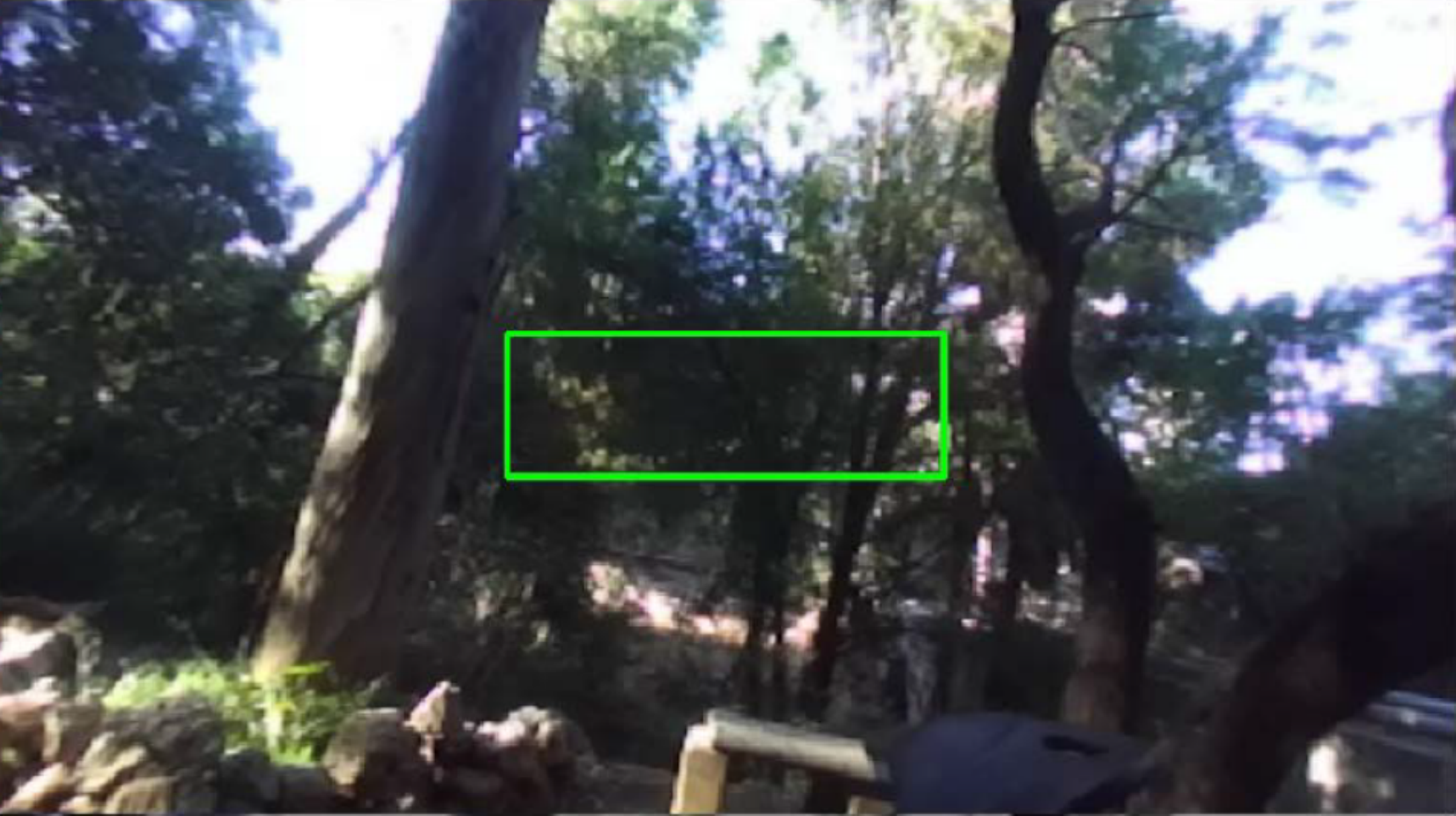}}
 \subfigure[No obstacle inside window in depth map.]{\includegraphics[width=40mm]{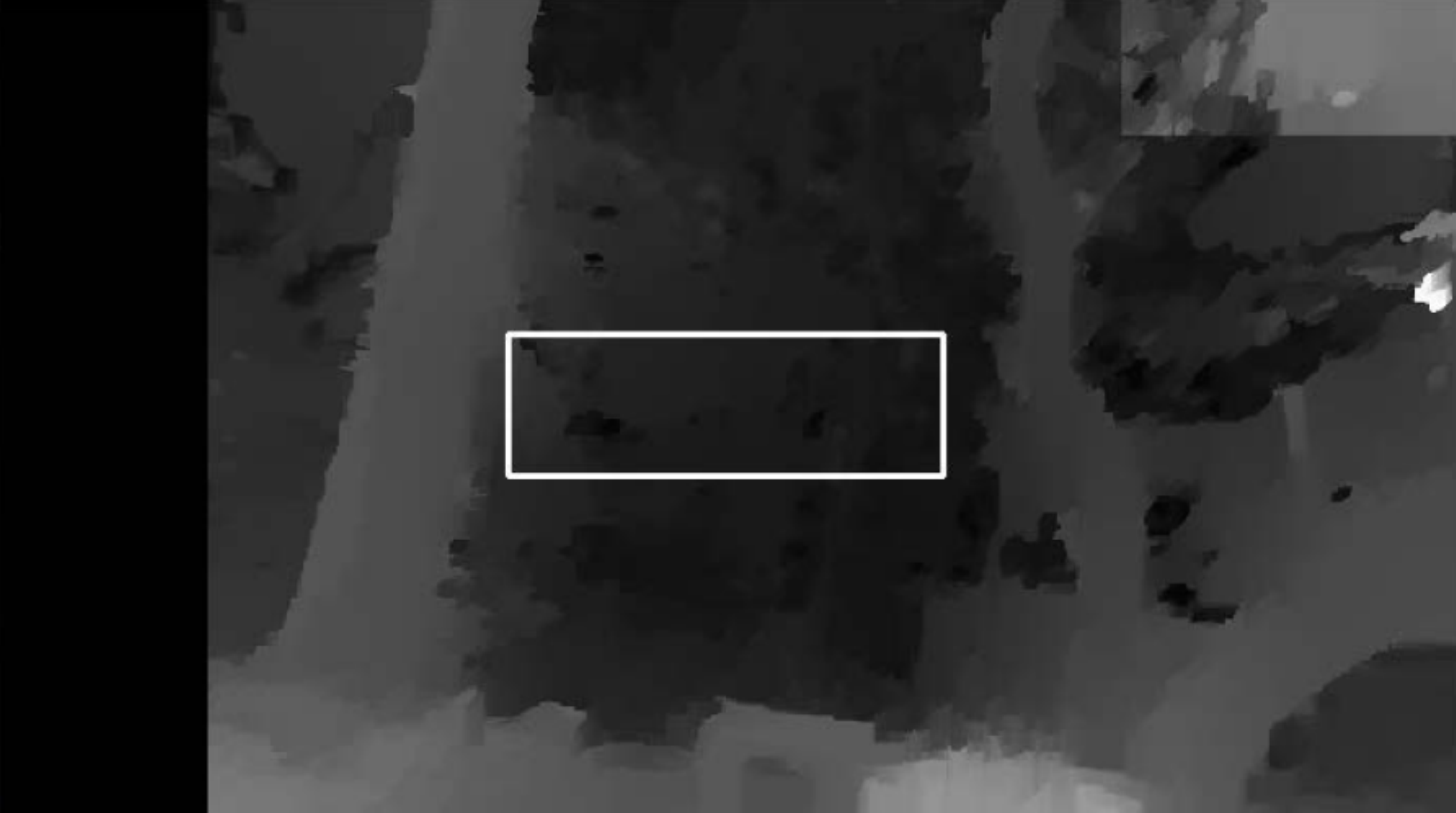}}
 \subfigure[Obstacle inside window in RGB]{\includegraphics[width=40mm]{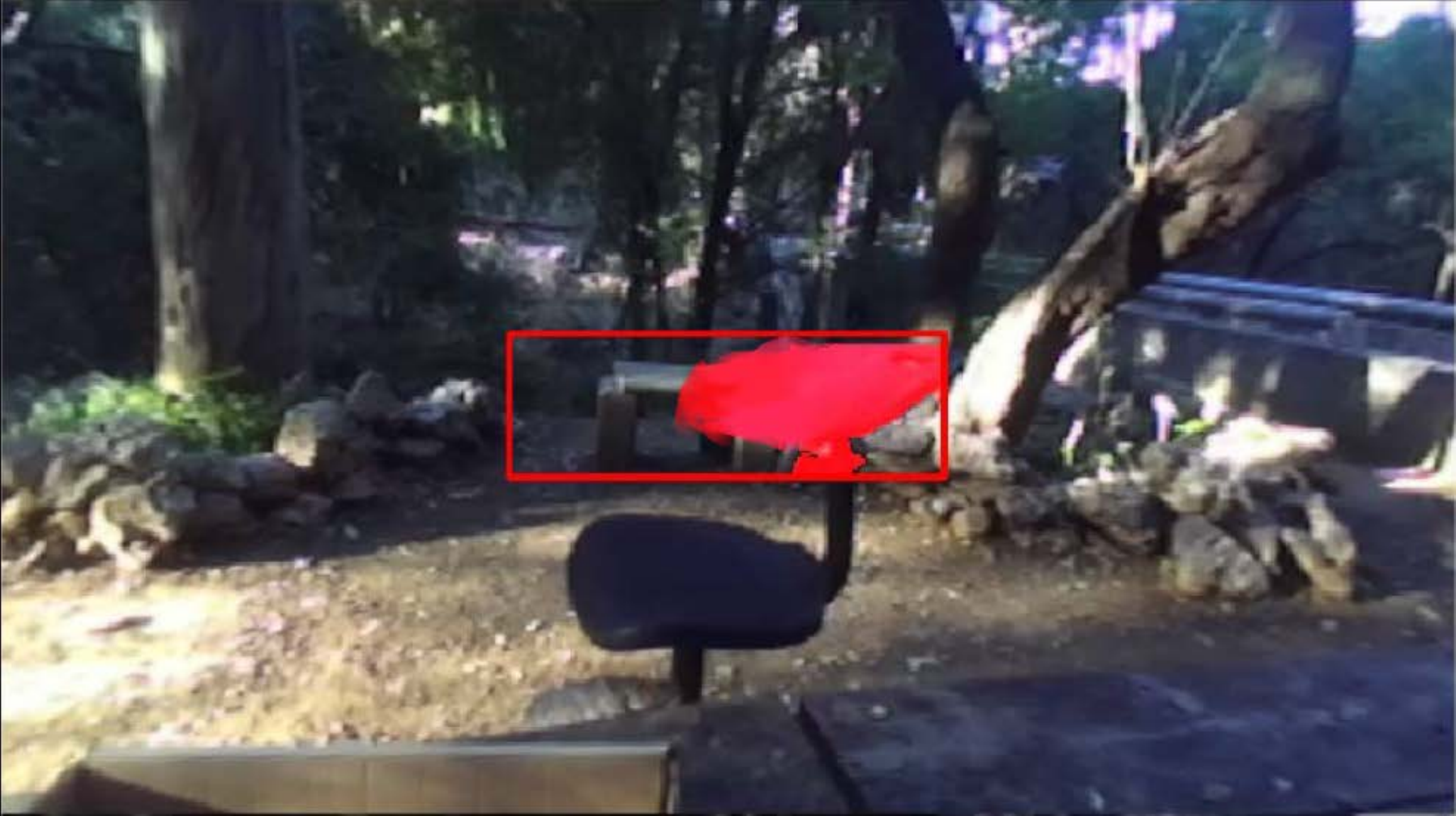}}
 \subfigure[Obstacle inside window in depth map]{\includegraphics[width=40mm]{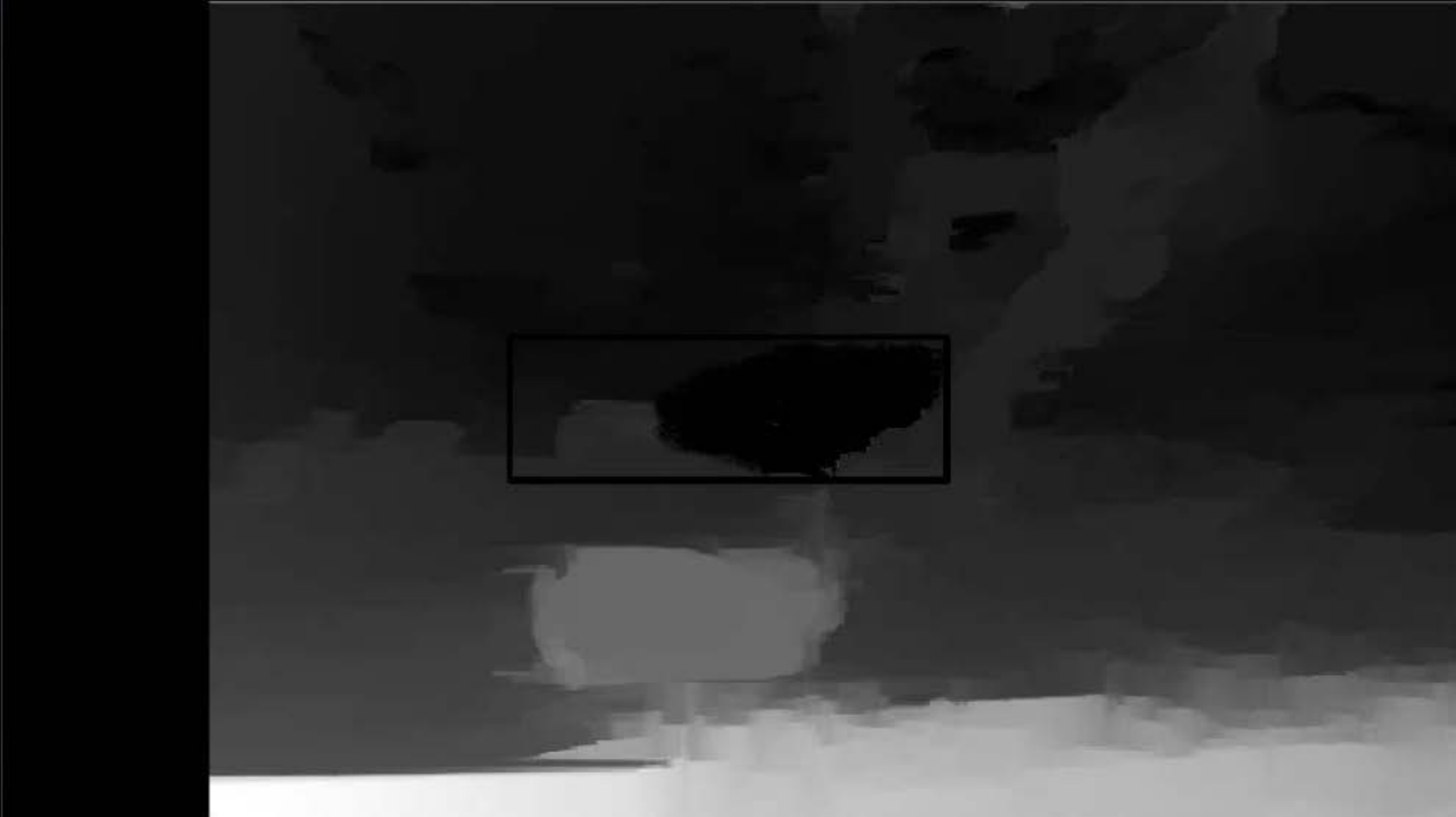}}
 \caption{Examples of detection of obstacle in exterior.}
 \label{exper_ext}
\end{figure}
In Fig. \ref{exper_ext} we show the results of obstacle detection in exterior when the detection window is 3 meters ahead from the quadrotor. In a) and b) the quadrotor is looking between two trees with no detection of obstacles. But if we put an obstacle inside the center and from 1.5 meters away, it is detected as an obstacle in that flying direction. 

\begin{figure}[t]
 \centering
    \subfigure[]{\includegraphics[width=\columnwidth]{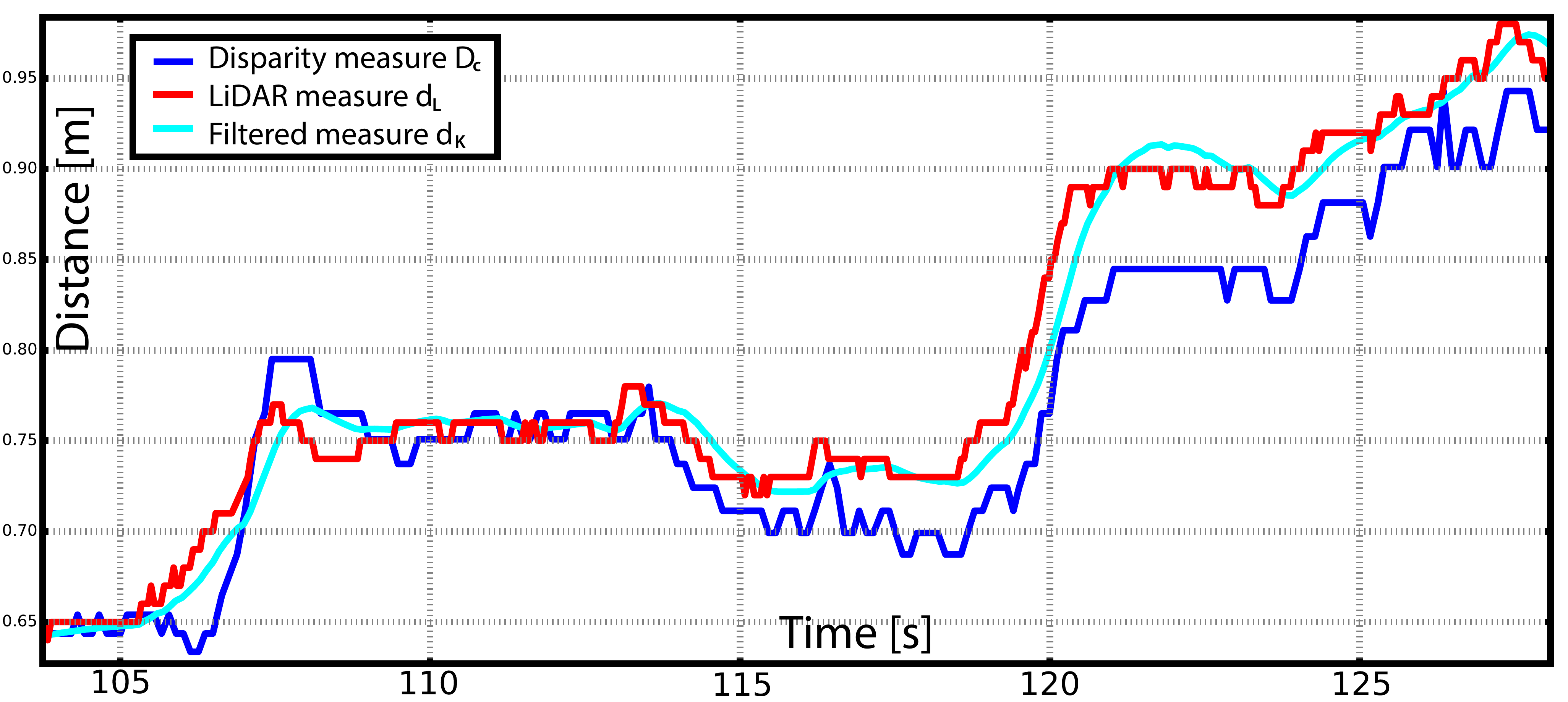}}
\subfigure[]{\includegraphics[width=\columnwidth]{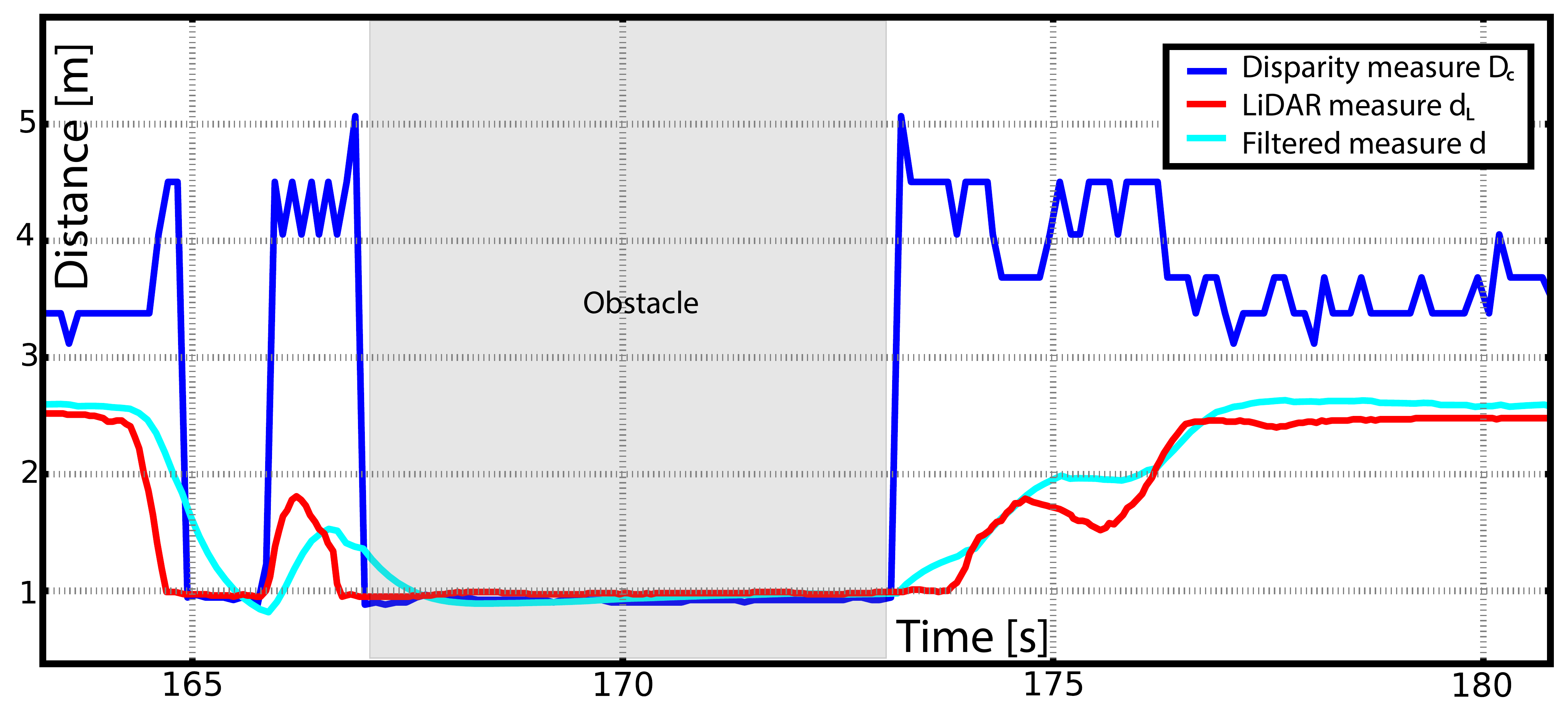}}
\subfigure[]{\includegraphics[width=\columnwidth]{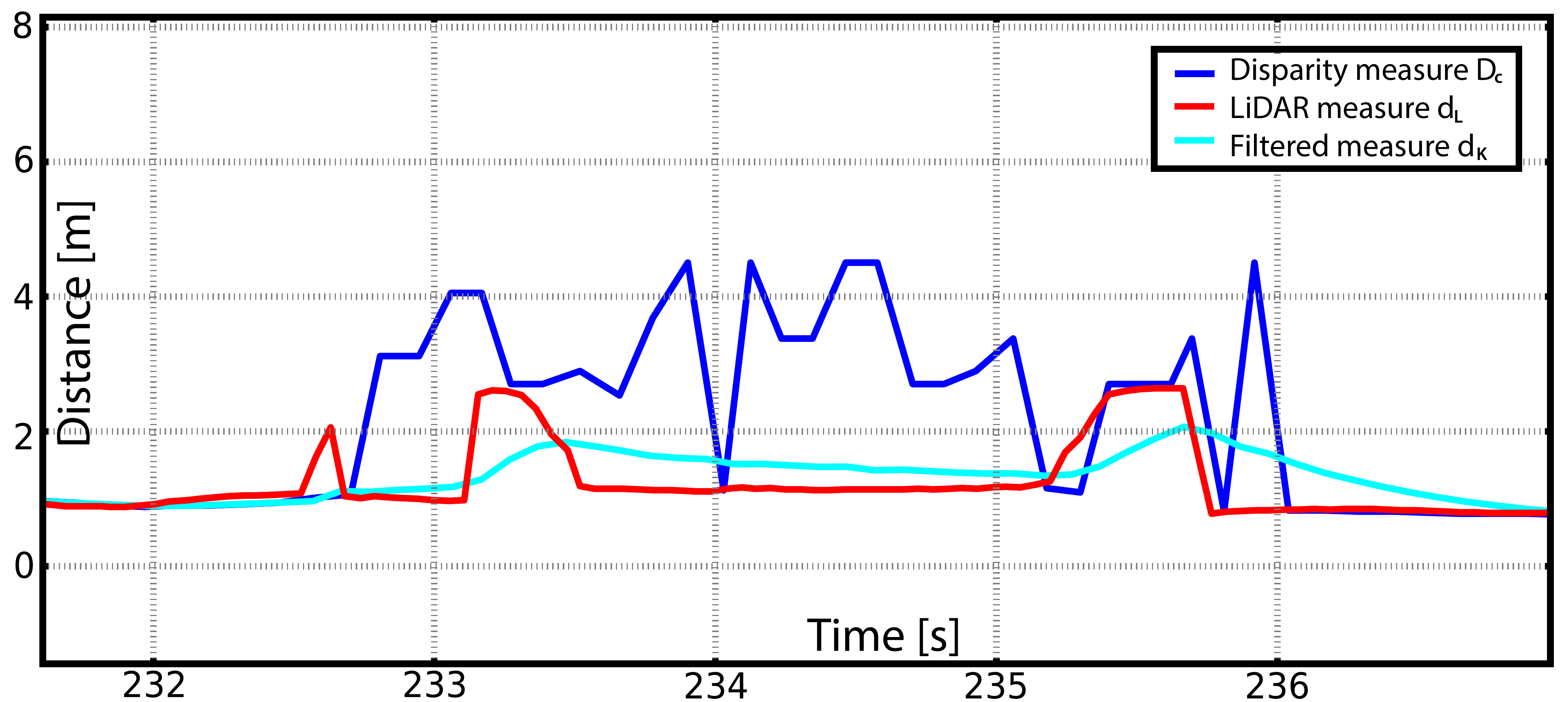}}
 \caption{Blue line represents the output of disparity, red line corresponds the output of 1D-LiDAR and cyan line shows the output from Kalman filter.}
 \label{KalmanGraph}
\end{figure}

Fig. \ref{KalmanGraph} shows the behavior of depth respect to disparity, 1D-LiDAR and Kalman Filter (KF) implementation; the distance is expressed in the $x$-axis in meters. The plot depicted at \ref{KalmanGraph} a) shows the change of depth signal when an obstacle is detected from both sensors. Also, depth obtained from ZED is noisier compared with the signal from 1D-LiDAR, so the covariance from ZED is bigger than covariance from 1D-LiDAR. Observe the effects of having a smaller 1D-LiDAR covariance than the disparity map coming from the stereo camera rig. At this point, one can pose the following question: What is the advantage of using disparity map instead of only 1D-LiDAR? The answer is that with the help of disparity map we can provide to the system information regarding visual perception. Also, we can extend the covered area of the potential obstacle; the 1D-LiDAR is used only to provide more precision to our estimation of free-obstacle area. At Fig. \ref{KalmanGraph} b), we can visualize what happens when the camera and 1D-LiDAR detect an obstacle. The depth from stereo camera increases but depth from sensor does not. Then, the peak caused by the computation of depth from stereo camera does not largely affect the filter response. On the other hand, We can observe at Fig. \ref{KalmanGraph} c) the behavior of the estimate depth map by KF when sensor provides peaks of different height in a very small instant. Kalman filter smooths the input signals, as we can verify in the image where the estimated signal given by the KF does not present peaks.
\section{Concluding remarks and future work}\label{sec:conc}
In this work we have presented a low-cost implementation to estimate a free navigation area in front of a quadrotor that is interpreted as a free-obstacle navigation area. For that aim, we propose a Kalman-filter-based algorithm that uses information from a stereo camera rig and a 1D-LiDAR. The results show that the estimation of free obstacle areas are considerably ameliorated with this approach since: a) using only a 1D-LiDAR the quadrotor cannot have any visual perception of their environment; and b) using only disparity maps conducts to a noise response. Then, mixing these senors in an appropriate Kalman Filter together with WLS filter results in more trustworthy depth map for obstacle navigation inside a depth window, which is a predefined area depending on the size of the drone in which this can navigate freely.

The implementation of the WLS filter greatly removes noise of the depth measuring, as we can see from Fig. \ref{KalmanGraph}, where it is shown that the depth map does not have any big peaks or NaN values. However, the distance becomes more imprecise when the distance increases and shows small peaks that represents the increase of a pixel in the disparity map. This is caused by the small distance between the pixel where is an object in the left image and the right image. So, while more depth, minus disparity and more noise. In this case the distance sensor does not suffer that problem. The graphs showed that the estimated measurement approaches to the laser sensor due to its variance was smaller than the depth map. In this way, we can compute with more precision the depth inside the depth window where is safe for the quadrotor to pass through.

As future work we intend to provide a stabilizer mechanism to the 1D-LiDAR and stereo camera rig in order to compensate aggressive tilting movements from the quadrotor. Also, the implementation of our approach in a navigation task with a navigation control algorithm is left as future work. Regarding computing time, we left as future work the implementation of our algorithm in GPU instead of only CPU.



\bibliographystyle{IEEEtran}
\bibliography{bib/References}

\end{document}